\documentclass[pmlr]{jmlr}

\usepackage{microtype}
\usepackage{booktabs}
\usepackage{colortbl}
\usepackage{wrapfig}
\usepackage{multirow}
\usepackage{xcolor}
\usepackage[font=small,labelfont=bf]{caption}
\usepackage[most]{tcolorbox}

\definecolor{navy}{HTML}{000080}
\hypersetup{
    colorlinks=true,
    citecolor=navy,
    linkcolor=navy,
    urlcolor=navy
}

\newcommand{\std}[1]{\ensuremath{{\scriptstyle\pm#1}}}

\definecolor{c4blue}{HTML}{346DFF}
\newtcolorbox{findingbox}{colback=c4blue!4, colframe=c4blue!55,
  boxrule=1.0pt, arc=1.5pt, left=3.5pt, right=3.5pt, top=2pt, bottom=2pt,
  before skip=2.5pt, after skip=2.5pt, fontupper=\small}
\newcommand{\finding}[2]{\begin{findingbox}\textbf{#1}~#2\end{findingbox}}

\jmlrvolume{}
\firstpageno{1}
\editors{}
\jmlryear{2026}
\jmlrworkshop{Symmetry and Geometry in Neural Representations}

\title[Benefit of Structured Representations]{Benefit of Structured Representations:
When Equivariance Helps in Image Classification}

\author{
   \Name{Florian A. Hölzl} \Email{florian.hoelzl@hpi.de}\\
   \addr Hasso Plattner Institute, Technical University of Munich (TUM)
   \AND
   \Name{Daniel Rueckert} \Email{daniel.rueckert@tum.de}\\
   \addr TUM and TUM University Hospital, Munich Center for Machine Learning, Imperial College London
   \AND
   \Name{Georgios Kaissis} \Email{georg.kaissis@hpi.de}\\
   \addr Hasso Plattner Institute
}

\begin{document}

\maketitle

\begin{abstract}
Equivariant neural networks promise improved representation learning through structural guarantees.
However, mainstream vision encoders rarely incorporate equivariance for most geometric symmetries.
Images are a particularly interesting modality with many such symmetries and extensive datasets from which to learn this structure.
In this work, we investigate when vision encoders benefit from geometric inductive biases.
We study the discrete rotation group of quarter-turns ($C_4$) which acts losslessly without interpolation or boundary artifacts.
Combined with image classification as a controlled invariant objective, this allows for a principled empirical investigation into the role of structural guarantees.
We introduce ConvNeXt-C4 as a rotation equivariant ConvNeXt with dimension-equivalent yet orientation-structured representations.
Training from scratch, our experiments show, that the $C_4$ variant outperforms its base model across all tasks while operating with 6$\times$ fewer parameters and 1.6$\times$ fewer FLOPs.
Its performance advantage increases log-linearly with decreasing training set size, with pre-training providing a similar improvement across both model variants.
Analyzing the classifier head reveals that orientation information from the representations' structure is actively used, improving performance by 4.90$\%$ across all datasets.
Our results highlight the sample-efficiency benefits of equivariance and the value of retained orientation information, motivating further investigation beyond image classification.
Code is available at \url{https://github.com/hlzl/equivariant-convnext}.
\end{abstract}

\section{Introduction}
\label{sec:intro}

Despite vast amounts of research on geometric deep learning in recent years, modern neural networks do not widely incorporate geometric inductive biases. 
This is despite the fact that the rigorous mathematical foundations of equivariant models for encoding geometric priors reduce hypothesis space complexity and provably bound generalization error~\citep{cohen2019theory, elesedy2023symmetry}.
Instead, learning these biases from massive amounts of pre-training data remains the predominant paradigm of most current approaches.
In this work, we ask: \textit{when do the beneficial properties of equivariant architectures actually benefit learning?}

Imaging is a particularly relevant area to study this question, with vast amount of data and research on equivariant methods available~\citep{cohen2016group, kondor2018general, thomas2018tensor, weiler2019general, satorras2021graph}.
The adoption gap of equivariant methods in imaging can be attributed to computational overhead and diminishing returns on modern architectures.
However, this trade-off has rarely been evaluated on modern backbones using systematic architectural interventions.
While recent geometric deep learning has shifted toward irregular modalities, 2D images represent the hardest benchmark as standard vision backbones are mature and highly optimized.
Image classification provides an ideal testbed where predictions should be invariant under rotation, ensuring any measured effect stems purely from the encoder.
Planar rotation is an exact symmetry for semantics in vision, with a rotated image depicting the same object.
It is thus a good starting point to better understand the role of geometric inductive biases in this competitive modality.

We postulate that evaluating geometric inductive biases in vision requires three key components.
First, a diverse group of training datasets with varying properties such as sample size and data symmetry.
Second, an equivariant architecture that is comparable to modern vision encoders in terms of efficiency and scalability.
Third, an analysis that clearly differentiates between (i) \emph{equivariant feature fields} that preserve information internally and (ii) \emph{invariant readout closure} that permanently eliminates information.

\begin{figure}[t]
  \centering
  \includegraphics[width=\linewidth]{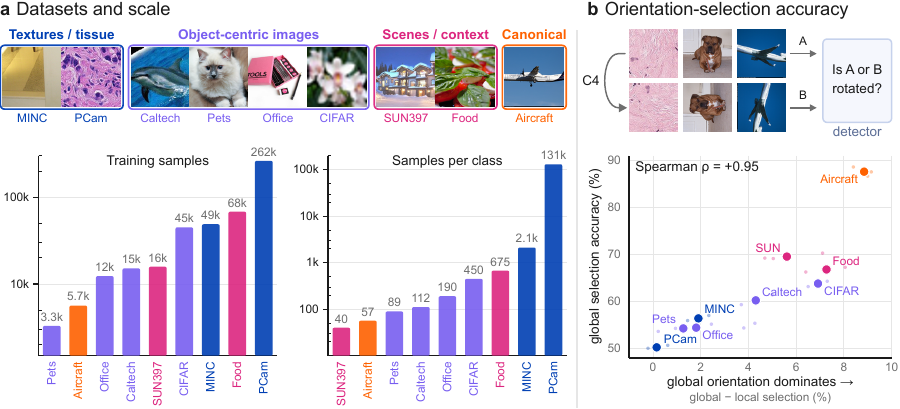}
  \caption{\textbf{Nine-dataset experimental panel spans image semantics, size and symmetry.}
  \textbf{(a)} Training samples span three orders of magnitude ($3.3\text{k}$ to $262\text{k}$).
  Different semantics and samples per class aid in assessing varying dataset challenges.
  \textbf{(b)} Determining orientation symmetry in data via classifier two-sample test (C2ST) accuracy shows how strongly global orientation dominates local patch symmetry (PCam most symmetric globally and locally; Aircraft strongest global orientation).}
  \label{fig:overview}
\end{figure}

In this work, we propose a $C_4$ rotation equivariant \textit{ConvNeXt}~\citep{liu2022convnext} closely resembling its base model.
We choose ConvNeXt because it is the only architecture proposed so far that rivals vision transformers in performance while maintaining strict translation equivariance~\citep{woo2023convnext, simeoni2025dinov3}.
The discrete rotation group of quarter-turns ($C_4$), which acts as an exact permutation of pixel coordinates on a square grid with zero interpolation or resampling artifacts, allows for isolating the effect of geometric symmetries.
Using an approach to factorize the equivariant convolution operation based on~\citet{bekkers2024ponita}, it remains a highly efficient architecture.
The model is evaluated across a benchmark panel of 9 vision datasets spanning three orders of magnitude in training set size alongside ImageNet-1K pre-training~\citep{deng2009imagenet}.
Our empirical findings reveal that $C_4$-equivariance substantially improves performance under data-scarcity and provides structure in the representation that is leveraged across all data regimes.

\section{Related Work}
\label{sec:related}

Increasing the inductive bias in vision encoders is one of the central early areas of geometric deep learning.
Focusing initially on rotations through regular group equivariance and steerable networks~\citep{cohen2016group, dieleman2016cyclic, weiler2019general}, research later extended to other affine transformations such as scale~\citep{worrall2019scale}.
While discretization artifacts on the equidistant pixel grid of images are an issue of equivariance, the area also benefits from a clear connection to generalization theory~\citep{cohen2019theory, elesedy2023symmetry}.
This theory is also closely connected to the strong performance of equivariance under training noise~\citep{hoelzl2023sparse, wang2025bridging}.

Recent geometric deep learning has either shifted toward more irregular modalities such as point clouds, graphs, and fields~\citep{satorras2021graph, bekkers2024ponita} or increasingly relaxed the tight equivariant constraint~\citep{romero2022partial, kaba2023canonicalization, vadgama2025probing}.
While interesting, relaxing equivariance makes it substantially more difficult to separate performance gains in the empirical setting.
Other recent works explore increasing geometric priors in the more general transformer architecture~\citep{bokman2025flopping, nordstrom2025octic}.
This is in line with works such as \citet{brehmer2025scale} that aim to analyze equivariance at scale.
To clearly separate the effects of different inductive biases, we focus on a convolutional architecture and on the discrete group $C_4$, providing lossless rotation and translation equivariance throughout the full encoder.

\section{Methods}
\label{sec:method}

\begin{figure}[t]
  \centering
  \includegraphics[width=\linewidth]{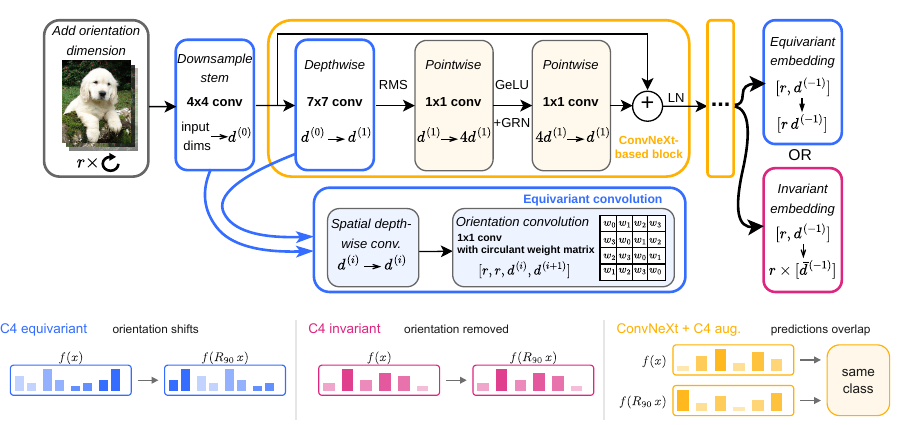}
  \caption{\textbf{Factorizing convolutions enables efficient equivariance with minimal adaptation.}
  Key architectural property of ConvNeXt-C4 is factorizing spatial depthwise convolutions with circulant orientation mixing. 
  This \textcolor[HTML]{346DFF}{preserves and permutes orientation features} instead of \textcolor[HTML]{DC267F}{eliminating orientation} information or learning rotated features with no formal geometric guarantees through \textcolor[HTML]{FFB000}{augmentations}.}
  \label{fig:architecture}
\end{figure}

\subsection{Exact $C_4$-Equivariant ConvNeXt}
\label{sec:architecture}

The planar rotation group $C_4 = \{0^\circ, 90^\circ, 180^\circ, 270^\circ\}$ acts losslessly on square pixel grids via exact permutation without interpolation artifacts. We design \textbf{ConvNeXt-C4} to match the stage depths $[3, 3, 27, 3]$ and total channel capacities $[96, 192, 384, 768]$ of ConvNeXt-Small \citep{liu2022convnext}.
Every intermediate feature map is structured as a regular-representation field over $C_4$ with shape $[\text{batch}, 4, C/4, H, W]$, where $4$ represents the orientations $\{0^\circ, \allowbreak 90^\circ, \allowbreak 180^\circ, \allowbreak 270^\circ\}$.
Constructing the representations in this scheme guarantees an equivalent embedding dimensionality between ConvNeXt and its equivariant counterpart.

Following \citet{bekkers2024ponita}, each residual block factorizes spatial and group operations (Figure~\ref{fig:architecture}, center).
The spatial depthwise convolution is a $7\times 7$ kernel shared and rotated across the 4 orientations.
Afterwards, we apply mixing across the orientation dimension by convolving with a circulant weight matrix.
This orientation convolution is effectively a $1\times 1$ regular-representation mixer that aggregates orientation channels:
\begin{equation}
    y[r_{\text{out}}] = \sum_{r_{\text{in}}=0}^3 W[(r_{\text{in}} - r_{\text{out}}) \bmod 4] \, x[r_{\text{in}}],
\end{equation}
where $W \in \mathbb{R}^{4 \times (C_{\text{out}}/4) \times (C_{\text{in}}/4)}$ learns all four relative rotational offsets.

The remaining pointwise convolutions only mix channel information and are equivariant with respect to our regular rotation group.
This also applies to the non-linearities and normalizations throughout the architecture.
Because parameters are shared across orientation planes, ConvNeXt-C4 operates with 6$\times$ fewer parameters and 1.6$\times$ fewer FLOPs relative to standard ConvNeXt (Table~\ref{tab:cost}).

\subsection{Equivariant vs.\ Invariant Readout Mechanisms}
\label{sec:readouts}

After global spatial pooling, the encoder yields 4 orientation vectors $\{z_r\}_{r=0}^3$, each of dimension $C/4 = 192$, following the $768$-dimensional output embedding of the base ConvNeXt-Small.
We evaluate two readout strategies before the final classification head:
\paragraph{Equivariant Feature Field (\textit{C4 equiv.}):} Concatenates orientation vectors directly, $z_{\text{eq}} = [z_0, z_1, z_2, z_3] \in \mathbb{R}^{768}$. 
An input rotation $R_g$ cyclically permutes these sub-vectors ($[z_r] \mapsto [z_{r+1}]$), preserving the geometric structure for downstream heads.
\paragraph{Invariant Representation (\textit{C4 inv.}):} Maps $\{z_r\}_{r=0}^3$ to the $C_4$ power spectrum, yielding an explicitly rotation-invariant descriptor $z_{\text{inv}} \in \mathbb{R}^{768}$ such that $f(R_g x) = f(x)$.
Unlike averaging, which retains only the zero-frequency component, the power spectrum preserves information from all $C_4$ Fourier modes.

\subsection{Disentangling Representation Structure: Head-Averaging Intervention}
\label{sec:head_intervention}

To determine whether downstream classifiers actively exploit the structured orientation information of an equivariant feature field in Section~\ref{sec:representation}, we perform a controlled weight-space intervention on trained classification heads.

A linear classifier maps the flattened representation $z_{\text{eq}} \in \mathbb{R}^{4 \times 192}$ to $K$ class logits using four learned weight blocks $[W_0, W_1, W_2, W_3] \in \mathbb{R}^{K \times 4 \times 192}$, one per orientation:
\begin{equation}
    \hat{y}_{\text{aware}} = \sum_{r=0}^3 W_r z_r.
\end{equation}
We construct an \textit{orientation-averaged} head by replacing each orientation block with their arithmetic mean $\bar{W} = \frac{1}{4}\sum_{r=0}^3 W_r$:
\begin{equation}
    \hat{y}_{\text{avg}} = \sum_{r=0}^3 \bar{W} z_r = \bar{W} \left( \sum_{r=0}^3 z_r \right).
\end{equation}
This intervention preserves each class's mean feature projection while eliminating orientation-specific weighting. The difference in accuracy ($\text{Acc}(\hat{y}_{\text{aware}}) - \text{Acc}(\hat{y}_{\text{avg}})$) isolates the performance contribution of retained geometric structure.

\subsection{Experimental Benchmark and Diagnostic Suite}
\label{sec:benchmark}

\paragraph{Dataset Panel and Training Protocol.}
We evaluate all models across nine benchmark classification datasets (Figure~\ref{fig:overview}a): Oxford Pets, FGVC-Aircraft, Office-Home, Caltech-256, SUN397, CIFAR-100, MINC-2500, Food-101, and PCam.
We compare four main model variants: ConvNeXt, ConvNeXt~+~$C_4$ aug., $C_4$ equivariant, and $C_4$ invariant.
All from-scratch models are trained at $224\times 224$ resolution with AdamW, batch size 256, and cosine schedules over three seeds.
For fine-tuning evaluations, we initialize the encoders after pre-training on ImageNet-1K followed by fine-tuning for 30 epochs.
More information on the datasets and the exact training setups are in Appendix~\ref{app:protocol}.

\paragraph{Dataset Symmetry Diagnostics.}
We ask whether datasets with greater rotational symmetry benefit more from equivariant encoders.
To quantify this symmetry, we use a classifier two-sample test (C2ST)~\citep{lopezpaz2017revisiting, lawrence2026augment}.
A binary classifier learns to distinguish unmodified images $x_i$ from randomly rotated images $R_g x_j$, where $x_i$ and $x_j$ are drawn from disjoint subsets of the dataset $X$ and $g$ is sampled uniformly from $C_4$.
On a balanced evaluation set, accuracy near $50\%$ indicates little detectable difference between the original and rotated images, whereas higher accuracy indicates stronger feature orientation.
The test is applied to whole images and randomly sampled $32^2$ patches to assess global and local rotational symmetry (Figure~\ref{fig:overview}b).
Further details are provided in Appendix~\ref{app:diagnostics}.

\section{Empirical Results}
\label{sec:results}

\subsection{From Scratch: Exact Equivariance Delivers Substantial Sample Efficiency}
\label{sec:scratch}

\begin{figure}[htbp]
  \centering
  \includegraphics[width=\linewidth]{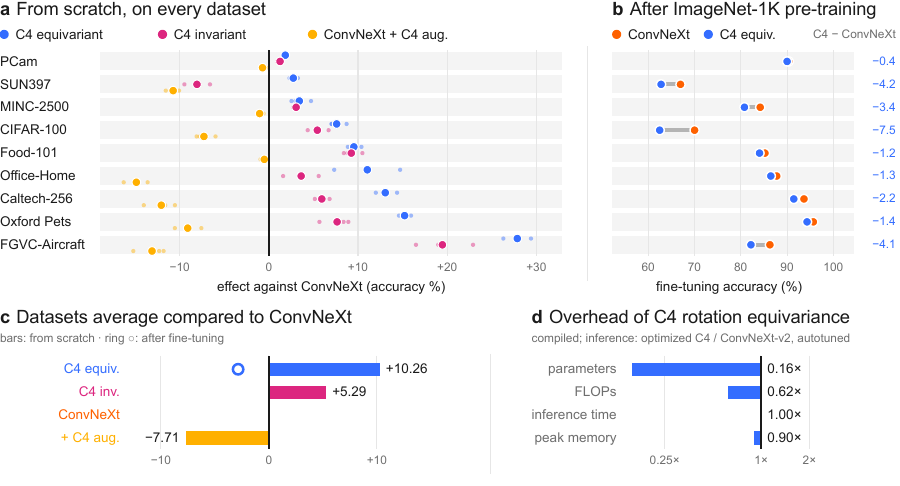}
  \caption{\textbf{Empirical performance across the nine-dataset panel.} 
  \textbf{(a)} Paired accuracy gains relative to baseline ConvNeXt across three seeds per dataset. 
  \textbf{(b)} Absolute accuracy achieved after ImageNet-1K pre-training and fine-tuning. 
  \textbf{(c)} Aggregate means across models (open ring indicates pre-trained $C_4$ equivariant mean). \textbf{(d)} Computational demands of $C_4$ (\textit{see Appendix~\ref{app:cost}}).}
  \label{fig:performance}
\end{figure}

Table~\ref{tab:panel} and Figure~\ref{fig:performance} present classification performance across all nine benchmark datasets.
Trained from scratch, ConvNeXt-C4 achieves a substantial performance advantage over baseline ConvNeXt. 
Across the nine datasets, ConvNeXt-C4 achieves a mean macro-accuracy improvement of \ensuremath{+10.26} points.
This gain is positive across all 9 dataset means and all 27 paired seeds, reaching up to \ensuremath{+27.87} points on fine-grained FGVC-Aircraft.

\begin{table}[htbp]
  \centering
  \footnotesize
  \setlength{\tabcolsep}{4pt}
  \begin{tabular}{llrrrrrrrrr}
\toprule
Init. & Model & \shortstack{Oxford\\Pets} & \shortstack{FGVC\\Aircraft} & \shortstack{Office\\Home} & \shortstack{Caltech\\256} & \shortstack{SUN397\\\vphantom{100}} & \shortstack{CIFAR\\100} & \shortstack{MINC\\2500} & \shortstack{Food\\101} & \shortstack{PCam\\\vphantom{100}} \\
\midrule
\multirow{4}{*}[-8.8pt]{Scratch} & \raisebox{2.7pt}{ConvNeXt} & \raisebox{-1pt}{\shortstack{56.13 \\ \std{1.22}}} & \raisebox{-1pt}{\shortstack{29.13 \\ \std{1.85}}} & \raisebox{-1pt}{\shortstack{54.72 \\ \std{1.77}}} & \raisebox{-1pt}{\shortstack{54.76 \\ \std{1.51}}} & \raisebox{-1pt}{\shortstack{35.79 \\ \std{0.42}}} & \raisebox{-1pt}{\shortstack{68.07 \\ \std{0.86}}} & \raisebox{-1pt}{\shortstack{64.13 \\ \std{0.09}}} & \raisebox{-1pt}{\shortstack{69.41 \\ \std{0.47}}} & \raisebox{-1pt}{\shortstack{83.18 \\ \std{0.42}}} \\
\addlinespace[3pt]
 & \raisebox{2.7pt}{+ C4 aug.} & \raisebox{-1pt}{\shortstack{47.03 \\ \std{0.27}}} & \raisebox{-1pt}{\shortstack{16.03 \\ \std{1.17}}} & \raisebox{-1pt}{\shortstack{39.86 \\ \std{1.24}}} & \raisebox{-1pt}{\shortstack{42.69 \\ \std{0.30}}} & \raisebox{-1pt}{\shortstack{25.05 \\ \std{0.96}}} & \raisebox{-1pt}{\shortstack{60.79 \\ \std{0.40}}} & \raisebox{-1pt}{\shortstack{63.10 \\ \std{0.30}}} & \raisebox{-1pt}{\shortstack{68.90 \\ \std{0.77}}} & \raisebox{-1pt}{\shortstack{82.47 \\ \std{0.63}}} \\
\arrayrulecolor{black!15}\cmidrule{2-11}\arrayrulecolor{black}
 & \raisebox{2.7pt}{C4 inv.} & \raisebox{-1pt}{\shortstack{63.78 \\ \std{1.18}}} & \raisebox{-1pt}{\shortstack{48.60 \\ \std{5.07}}} & \raisebox{-1pt}{\shortstack{58.35 \\ \std{0.36}}} & \raisebox{-1pt}{\shortstack{60.70 \\ \std{0.81}}} & \raisebox{-1pt}{\shortstack{27.72 \\ \std{1.81}}} & \raisebox{-1pt}{\shortstack{73.51 \\ \std{0.41}}} & \raisebox{-1pt}{\shortstack{67.19 \\ \std{0.38}}} & \raisebox{-1pt}{\shortstack{78.66 \\ \std{0.68}}} & \raisebox{-1pt}{\shortstack{84.43 \\ \std{0.41}}} \\
\addlinespace[3pt]
 & \raisebox{2.7pt}{C4 equiv.} & \textbf{\raisebox{-1pt}{\shortstack{71.35 \\ \std{1.18}}}} & \textbf{\raisebox{-1pt}{\shortstack{57.00 \\ \std{2.95}}}} & \textbf{\raisebox{-1pt}{\shortstack{65.77 \\ \std{1.97}}}} & \textbf{\raisebox{-1pt}{\shortstack{67.82 \\ \std{1.27}}}} & \textbf{\raisebox{-1pt}{\shortstack{38.53 \\ \std{0.27}}}} & \textbf{\raisebox{-1pt}{\shortstack{75.67 \\ \std{0.14}}}} & \textbf{\raisebox{-1pt}{\shortstack{67.55 \\ \std{1.09}}}} & \textbf{\raisebox{-1pt}{\shortstack{78.95 \\ \std{0.34}}}} & \textbf{\raisebox{-1pt}{\shortstack{85.02 \\ \std{0.61}}}} \\
\arrayrulecolor{black!15}\midrule\arrayrulecolor{black}
\multirow{2}{*}[-2.6pt]{IN-1K} & \raisebox{-1pt}{ConvNeXt} & \textbf{\raisebox{-1pt}{95.68}} & \textbf{\raisebox{-1pt}{86.30}} & \textbf{\raisebox{-1pt}{87.77}} & \textbf{\raisebox{-1pt}{93.66}} & \textbf{\raisebox{-1pt}{66.99}} & \textbf{\raisebox{-1pt}{70.03}} & \textbf{\raisebox{-1pt}{84.21}} & \textbf{\raisebox{-1pt}{85.23}} & \textbf{\raisebox{-1pt}{90.37}} \\
\addlinespace[3pt]
 & \raisebox{2.7pt}{C4 equiv.} & \raisebox{-1pt}{\shortstack{94.32 \\ \std{0.47}}} & \raisebox{-1pt}{\shortstack{82.20 \\ \std{0.30}}} & \raisebox{-1pt}{\shortstack{86.50 \\ \std{0.19}}} & \raisebox{-1pt}{\shortstack{91.46 \\ \std{0.04}}} & \raisebox{-1pt}{\shortstack{62.80 \\ \std{0.41}}} & \raisebox{-1pt}{\shortstack{62.49 \\ \std{0.23}}} & \raisebox{-1pt}{\shortstack{80.80 \\ \std{0.21}}} & \raisebox{-1pt}{\shortstack{84.06 \\ \std{0.02}}} & \raisebox{-1pt}{\shortstack{90.01 \\ \std{0.70}}} \\
\bottomrule
\end{tabular}

  \caption{Validation accuracy in $\%$ (mean $\pm$ sample standard deviation across three seeds).}
  \label{tab:panel}
\end{table}

Comparing the three operational interventions reveals a consistent hierarchical accuracy ordering across 8 of the 9 datasets:
$\text{C4 Equivariant} > \allowbreak \text{C4 Invariant} > \allowbreak \text{ConvNeXt} > \allowbreak\text{ConvNeXt + C4 Aug.}$
First, the equivariant readout outperforms the invariant readout by \ensuremath{+4.97} points across all nine datasets at identical capacity, proving that retaining orientation information is strictly advantageous over invariant collapse.
Second, enforcing invariance architecturally via the invariant readout outperforms teaching invariance via rotation augmentation by \ensuremath{+13.00} points across all nine datasets.
Third, adding exact $C_4$ rotation augmentation to standard ConvNeXt actively degrades accuracy by \ensuremath{-7.71} points across all datasets and seeds.
The performance advantage of $C_4$ Invariant architectures is remarkable and points to the fact that preserving orientation information throughout the network seems preferable to presenting the network random augmentations and expecting it to disentangle orientation information during training. 

\paragraph{Ruling out Confounders}
To rule out that the enhanced inductive bias of equivariant architectures causes memorization rather than improving generalization, we compared ConvNeXt-C4 to ConvNeXt under distribution shifts, confirming that gains reflect genuine representation learning:
ConvNeXt-C4 maintains a lead on duplicate-free ciFAIR-100~\citep{barz2020cifair} and on three natural shift targets (Pets $\to$ Dogs by \ensuremath{+15.26}, PCam $\to$ Camelyon17~\citep{koh2021wilds} by \ensuremath{+2.25}, SUN397 $\to$ Places365~\citep{zhou2017places} by \ensuremath{+1.49}).
See Appendix~\ref{app:shift} for further details.
Moreover, the success of ConvNeXt-C4 is not a capacity or regularization artifact:
(1) equivariant and invariant readouts share identical parameters (7.76M) yet differ by \ensuremath{+4.97} points;
(2) explicit $C_4$-rotation augmentation degrades performance by \ensuremath{-7.71} points;
and (3) switching out the architecture for the much smaller ConvNeXt Atto diminishes the equivariant gain ($+4.80$ vs.\ $+11.84$ points), hinting at potential further benefits when increasing the size of our $C_4$-variants.
\vspace{1em}
\finding{Finding 1: Exact rotation equivariance delivers substantial sample efficiency at reduced parameter and compute cost.}{Trained from scratch, ConvNeXt-C4 outperforms standard ConvNeXt by \ensuremath{+10.26} macro-accuracy points across all datasets and seeds while utilizing 6$\times$ fewer parameters and 10\% less training memory.
Retaining orientation information consistently outperforms invariant readout mappings and rotation data augmentation.}

\begin{figure}[tbp]
  \centering
  \includegraphics[width=\linewidth]{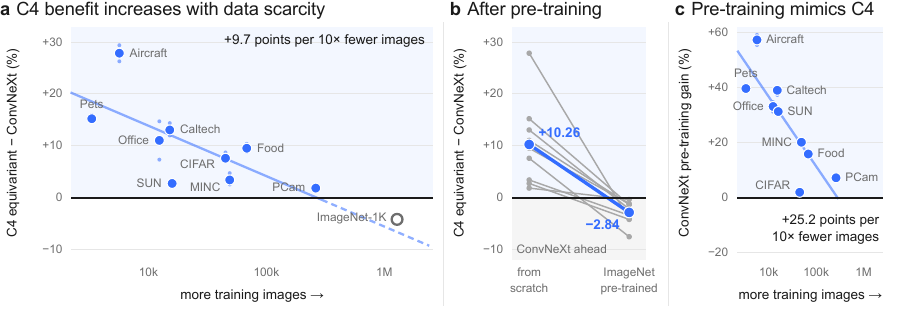}
  \caption{\textbf{The data-regime dependency of geometric inductive bias.} \textbf{(a)} From-scratch macro-accuracy gain of ConvNeXt-C4 over ConvNeXt plotted against training set size $N$ (log-scale), showing a strong log-linear decay ($\rho=\ensuremath{-0.82}, p=0.011$). Open marker indicates ImageNet-1K top-1 gap (unfitted).
  \textbf{(b)} Paired dataset performance before and after ImageNet-1K pre-training (blue:average).
  \textbf{(c)} Macro-accuracy gain from ImageNet-1K pre-training on standard ConvNeXt on the same axis.
  }
  \label{fig:regime}
\end{figure}

\subsection{Data-Regime Dependency}
\label{sec:regime}
Across the nine datasets, the from-scratch gain of ConvNeXt-C4 over ConvNeXt decays log-linearly with data scale (Figure~\ref{fig:regime}a):
$\Delta \text{Acc}(N) = -9.72 \log_{10}(N) + 52.74$ ($\rho = \ensuremath{-0.82}, p = 0.011$).
The gain drops by 9.72 points per tenfold increase in training set size, predicting break-even at $N \approx 267{,}000$ images ($\approx11\times$ data multiplier).
We see similar results for transformer-based architectures in Appendix~\ref{app:vit}.
Following ImageNet-1K pre-training, the architectural gap reverses from \ensuremath{+10.26} points from scratch to \ensuremath{-2.84} points after fine-tuning (Figure~\ref{fig:regime}b).
Crucially, datasets where equivariance helps most from scratch are precisely those where pre-training helps unconstrained ConvNeXt most (Spearman $\rho = \ensuremath{+0.80}, p = 0.014$) (Figure~\ref{fig:regime}c).
In other words, equivariance compensates exactly those aspects of pre-training that are hardest to learn from existing data sources for unconstrained ConvNeXt.

Strikingly, the existence of symmetry on a dataset level does \textit{not} benefit equivariant architectures the most: input orientation-selection accuracy (C2ST) exhibits no correlation with from-scratch gains ($|\rho| < 0.50$).
In fact, the FGVC-Aircraft dataset that exhibits the least amount of intra-dataset symmetries (C2ST score 0.876) yields the largest gain (\ensuremath{+27.87} points), whereas PCam, a dataset of histopathology slides in which orientation plays no role at all (C2ST score 0.502) yields only \ensuremath{+1.85} points.
This is particularly interesting, as works in the field often focus on rotated datasets or modalities without native orientation (see e.g., \citet{dieleman2016cyclic,veeling2018pathology}).
\vspace{1em}
\finding{Finding 2: Equivariance compensates for data scarcity, not for the existence of intra-dataset symmetries.}{Accuracy advantage of equivariance is log-linear as training set size shrinks, making $C_4$ structure a multiplier for labeled data.
Pre-training can substitute for this sample efficiency, while intra-dataset visual symmetry fails to predict where equivariance helps.}

\subsection{The importance of orientation information for the classifier}
\label{sec:representation}
As described in Section~\ref{sec:head_intervention}, we next ask: does the fact that pre-training compensates or outdoes equivariance imply that structured representations are no longer useful? 
We thus perform a controlled weight-space intervention on trained ConvNeXt-C4 classification heads. 
The head maps $768$-D flattened feature vectors $[z_r]_{r=0}^3 \in \mathbb{R}^{4 \times 192}$ to class logits via weight blocks $[W_r]_{r=0}^3 \in \mathbb{R}^{K \times 4 \times 192}$. 
We construct an \emph{orientation-averaged} head by replacing each block with the mean matrix $\bar{W} = \frac{1}{4}\sum_{r=0}^3 W_r$, preserving each class's mean feature projection while destroying orientation-specific weighting. 
The accuracy of the trained \emph{orientation-aware} head over this aggregated copy is the \emph{orientation gain}.

\begin{figure}[htbp]
  \centering
  \includegraphics[width=\linewidth]{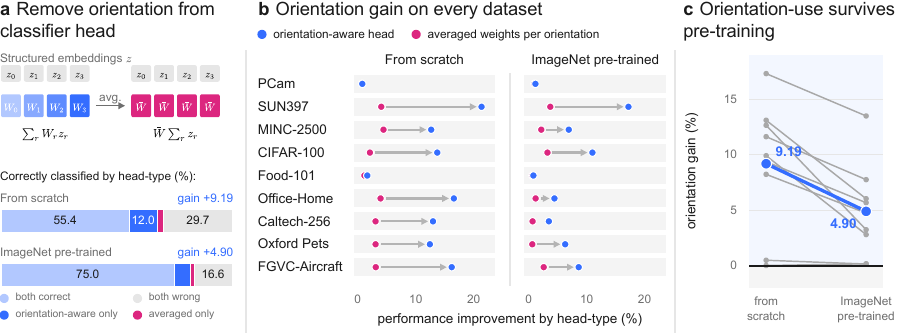}
  \caption{\textbf{Downstream classifiers benefit from retained orientation information.}
  \textbf{(a)} Head block-averaging intervention: replacing the four learned orientation weight blocks $W_r$ with their uniform average $\bar{W}$.
  The embedding blocks $z_r$ are untouched, so the edited head can read only their sum.
  \textbf{(b)} Per dataset, the accuracy each head type contributes beyond the samples both classify correctly, i.e.\ the share it alone gets right; arrows indicate orientation gain.
  \textbf{(c)} The orientation gain across from-scratch and pre-trained regimes.
  All percentages are macro-averaged over classes.}
  \label{fig:orientation}
\end{figure}

Figure~\ref{fig:orientation} details these findings:
from scratch, the orientation-aware head is uniquely correct on 12.03 points versus 2.84 points for the averaged head, an orientation gain of 9.19 points.
Following ImageNet-1K pre-training, the proportion of samples correctly classified by both heads rises from 55.4\% to 75.0\%, reflecting a ``rising tide" phenomenon where stronger generic features help both architectures. 
Yet, the orientation-aware head remains uniquely correct on 6.67 points versus 1.77 points for the averaged head, maintaining a 4.90 point gain.
This orientation gain remains strictly positive across all datasets in both regimes.
In other words, while pre-training moderately reduces reliance on orientation (reduction of 4.29 points, $p=0.008$), downstream classifiers consistently benefit from the retained geometric information.
On the datasets with the most symmetry (e.g., PCam), the differences in performance are expectedly minimal.

\vspace{1em}
\finding{Finding 3: Downstream classifiers leverage retained orientation information.}{Across all nine datasets, this yields an orientation gain of 9.19 points from scratch and 4.90 points after pre-training over an orientation-averaged baseline. While pre-training strengthens generic invariant representations, classifiers continue to actively exploit structured information.}

\section{Discussion}
\label{sec:discussion}

\paragraph{Design Guidelines for Practitioners.}
Based on our empirical findings, we distill four concrete guidelines for incorporating geometric inductive biases into vision encoders:\newline
(1)~\textbf{Budget Structure Like Capacity:} Treat exact equivariance as a sample-efficiency intervention. Introduce exact equivariant architectures when labeled training data is scarce ($N < 10^5$). 
When abundant data or web-scale pre-training is available, unconstrained architectures can achieve equivalent or superior classification accuracy.\newline
(2)~\textbf{Preserve Equivariant Feature Fields:} Retain explicit transformation information throughout the network and export equivariant feature fields to downstream heads. 
Forcing early invariant closure eliminates geometric information that classifiers actively exploit.\newline
(3)~\textbf{Avoid Synthetic Invariance via Heavy Augmentation:} Enforcing rotational invariance through data augmentation in capacity-limited networks incurs optimization interference and consistently degrades performance. 
If invariance is strictly required, enforce it architecturally via invariant readouts.\newline
(4)~\textbf{Do Not Rely on Input-Level Visual Symmetry:} The utility of rotational equivariance is determined by training set scale rather than whether images appear visually isotropic to human inspection.

\paragraph{Implications for Foundation Models and Geometric Representations.}
Our results offer a unifying perspective on the role of geometric representations. 
Scalar image classification represents a worst-case scenario for demonstrating the value of equivariant feature fields: because the target label is invariant, strong unconstrained foundation models can solve the task by learning invariant global descriptors. 
Yet, even in this unfavorable setting, our head-intervention experiments demonstrate that classifiers actively rely on the structured orientation information. 
This suggests that for tasks requiring geometric preservation---such as dense spatial prediction, 3D pose estimation, robotics, and generative modeling---structured equivariant representations provide foundational inductive biases that scalar classification benchmarks systematically undervalue.

\paragraph{Limitations and Future Work.}
Our study intentionally focuses on exact $C_4$ rotations on a single modern convolutional backbone family (ConvNeXt-Small) in single-view image classification to isolate mechanisms without confounding variables.
Investigating the benefit of geometric inductive biases in more challenging tasks such as vision-language modeling or diffusion image generation, where initial promising results have recently been demonstrated \citep{singh2026what, kouzelis2025eqvae}, represent fruitful directions for future research.

\acks{
The project on which this publication is based was funded by the Federal Ministry of Research, Technology and Space under the funding code “KI-Servicezentrum Berlin-Brandenburg” 16IS22092.
Responsibility for the content of this publication remains with the authors.
}

\bibliography{references}

\appendix

\newpage
\section{Implementation Details}\label{app:protocol}

\subsection{Hyperparameters and Training Configurations}

\paragraph{Training from Scratch.}
\begin{table}[hbpt]
  \centering
  \defcitealias{krizhevsky2009cifar}{CIFAR-100}
  \defcitealias{veeling2018pathology}{PCam}
  \defcitealias{bell2015minc}{MINC-2500}
  \defcitealias{griffin2007caltech}{Caltech-256}
  \defcitealias{bossard2014food}{Food-101}
  \defcitealias{venkateswara2017officehome}{Office-Home}
  \defcitealias{maji2013aircraft}{FGVC-Aircraft}
  \defcitealias{xiao2010sun}{SUN397}
  \defcitealias{parkhi2012pets}{Oxford Pets}
  \defcitealias{deng2009imagenet}{ImageNet-1K}
  
  \caption{Hyperparameters with peak learning rates given as ConvNeXt/C4 values.}
   \label{tab:training}
  \small
  \setlength{\tabcolsep}{3pt}
  \begin{tabular}{@{}lrrrrrrl@{}}
    \toprule
    Dataset & Image size & Epochs & Peak LR & Warm-up &
    Weight decay & Drop path & Crop \\
    \midrule
    \citetalias{krizhevsky2009cifar}     & $224^2$  & 150 & $3\times10^{-4}/10^{-3}$ & 5  & 0.05 & 0.4 & Padded \\
    \citetalias{veeling2018pathology}          & $96^2$  & 30  & $3\times10^{-4}/10^{-3}$ & 5  & 0.05 & 0.4 & Padded \\
    \citetalias{bell2015minc}     & $224^2$ & 90  & $3\times10^{-4}/10^{-3}$ & 5  & 0.05 & 0.4 & $[0.08,1.0]$ \\
    \citetalias{griffin2007caltech}   & $224^2$ & 300 & $3\times10^{-4}/10^{-3}$ & 5  & 0.05 & 0.4 & $[0.08,1.0]$ \\
    \citetalias{bossard2014food}      & $224^2$ & 150 & $3\times10^{-4}/10^{-3}$ & 5  & 0.05 & 0.4 & $[0.08,1.0]$ \\
    \citetalias{venkateswara2017officehome}{}   & $224^2$ & 90  & $3\times10^{-4}/10^{-3}$ & 5  & 0.05 & 0.4 & $[0.08,1.0]$ \\
    \citetalias{maji2013aircraft} & $224^2$ & 300 & $3\times10^{-4}/10^{-3}$ & 5  & 0.10 & 0.4 & $[0.25,1.0]$ \\
    \citetalias{xiao2010sun}        & $224^2$ & 150 & $3\times10^{-4}/10^{-3}$ & 5  & 0.05 & 0.4 & $[0.08,1.0]$ \\
    \citetalias{parkhi2012pets}   & $224^2$ & 300 & $3\times10^{-4}/10^{-3}$ & 5  & 0.10 & 0.4 & $[0.08,1.0]$ \\
    \citetalias{deng2009imagenet}   & $224^2$ & 150 & $10^{-3}/3\times10^{-3}$ & 10 & 0.05 & 0   & $[0.08,1.0]$ \\
    \bottomrule
  \end{tabular}
\end{table}

All datasets utilize fixed, class-stratified splits.
AdamW uses $\beta=(0.9,0.999)$ in all stages. Cosine decay generally ends at zero and $10^{-6}$ on ImageNet-1K.
ImageNet pre-training uses RandAugment magnitude 9, color jitter 0.4, random erasing 0.25, Mixup 0.8, CutMix 1.0, EMA decay 0.99995, and gradient clipping at 1.0.
Fine-tuning loads the EMA encoder and reinitializes its classifier with standard deviation 0.02.
"Padded" denotes a random crop with four-pixel padding.
All downstream arms use horizontal flips, PCam also uses vertical flips, and only the rotation augmentation controls add quarter-turn or $\pm30^\circ$ rotations.
We generally found that ConvNeXt-C4 needs a peak learning rate that is about $3\times$ higher than for the equivalent ConvNeXt.

\paragraph{Fine-Tuning Configuration.}
In comparison to training from scratch, fine-tuning is done for 30 epochs across all datasets.
Warmup is reduced to 1 epoch, drop path rate is set to 0.0, weight decay is consistently at 0.05, and crop scale at $[0.5, 1.0]$.
The image size for CIFAR-100 is increased to $224^2$.
The classification head is re-initialized randomly with a fixed seed.

\subsection{Computation Cost Analysis}\label{app:cost}

We compare the computation costs of both ConvNeXt-Small and our corresponding ConvNeXt-C4-Small in Table~\ref{tab:cost}.
ConvNeXt-C4 has $6\times$ fewer parameters and 37\% fewer counted FLOPs, yet these reductions do not translate directly into faster inference.
Without further improvements, the implementation leads to many small batched matrix products compared to the non-equivariant model and an increase in inference time by more than $2\times$.
This issue has already been described in previous works such as \cite{zhao2026flash}.
Their linear layer implementation does further reduce FLOPs yet did not provide throughput benefits for our model where memory traffic and embedding reshapes are the key limiting factors.
Alternative work on leveraging equivariance for efficiency using reflection-even/odd decomposition~\citep{bokman2025flopping} does not directly apply to our $C_4$ case, as rotation action contains coupled two-dimensional real components.

\begin{wraptable}{l}{0.5\textwidth}
  \centering
  \caption{Computational cost benchmark at $224^2$ image resolution, batch size 256, BF16 precision, compiled execution on an NVIDIA RTX A6000 GPU for \textit{small} model sizes.}
  \label{tab:cost}
  \small
  \resizebox{\linewidth}{!}{\begin{tabular}{@{}lrr@{}}
\toprule
Metric & ConvNeXt & C4 \\
\midrule
Parameters (M) & 50.32 & 8.44 \\
FLOPs (G) & 17.37 & 10.85 \\
Train (img/s) & 666 & 696 \\
Train memory (GiB) & 20.19 & 19.92 \\
\midrule
Inference (img/s) & & \\
\hspace{1em}Initial & 2321 & 1089 \\
\hspace{1em}+ Cache, reshape, merge & 2321 & 2258 \\
\hspace{1em}+ Autotuning & 2407 & 2411 \\
\bottomrule
\end{tabular}
}
\end{wraptable}

While most FLOP savings in our architecture come from the smaller MLP projection, the embedding size remains unchanged and thus no corresponding improvements with regard to normalization.
Instead, layout changes of these embeddings contribute to a runtime overhead of the $C_4$ model.
A key intervention to reduce this overhead is reshaping the equivariant embeddings into large two-dimensional matrices.
This reduces $C_4$ batch time by more than half without changing the underlying mathematical operations.
Caching weights and merging the stem and downsampling convolutions at inference removes further overhead.
With default PyTorch compilation, $C_4$ takes 113.2 ms per batch versus 109.7 ms for ConvNeXt.
Applying compiler autotuning to both models, with CUDA graphs disabled, brings their batch times to 107.3 and 107.1 ms, respectively.
$C_4$ therefore reaches throughput parity at this operating point while retaining its parameter and arithmetic savings.

\FloatBarrier
\section{Additional Experimental Results}

\subsection{Full Benchmark Panel and Continuous Rotation Analysis}\label{app:panel}

\begin{table}[hbpt]
  \centering
  \caption{Validation class-macro accuracy (mean $\pm$ sample standard deviation across three seeds) for all seven model arms, ordered by training set size. Column headers: \mbox{+C4} and \mbox{+30} denote ConvNeXt with exact $C_4$ and continuous $\pm 30^\circ$ augmentation; \mbox{C4-eq} and \mbox{C4-inv} denote ConvNeXt-C4 with equivariant and invariant readouts. Bold denotes the highest performing arm per dataset.}
  \label{tab:panel_full}
  \footnotesize
  \setlength{\tabcolsep}{3pt}
  \begin{tabular}{lrrrrrrr}
\toprule
Dataset & ConvNeXt & +C4 & +30 & C4-eq & C4-eq+30 & C4-inv & C4-inv+30 \\
\midrule
Oxford Pets & 56.13 \std{1.22} & 47.03 \std{0.27} & 55.86 \std{1.22} & \underline{71.35} \std{1.18} & \textbf{71.62 \std{1.40}} & 63.78 \std{1.18} & 61.44 \std{1.74} \\
FGVC-Aircraft & 29.13 \std{1.85} & 16.03 \std{1.17} & 25.67 \std{1.74} & \textbf{57.00 \std{2.95}} & \underline{49.17} \std{2.71} & 48.60 \std{5.07} & 36.50 \std{6.03} \\
Office-Home & 54.72 \std{1.77} & 39.86 \std{1.24} & 51.98 \std{0.70} & \textbf{65.77 \std{1.97}} & \underline{65.39} \std{0.97} & 58.35 \std{0.36} & 56.16 \std{0.29} \\
Caltech-256 & 54.76 \std{1.51} & 42.69 \std{0.30} & 50.04 \std{0.51} & \textbf{67.82 \std{1.27}} & \underline{67.60} \std{0.67} & 60.70 \std{0.81} & 57.28 \std{0.18} \\
SUN397 & \underline{35.79} \std{0.42} & 25.05 \std{0.96} & 32.09 \std{0.16} & \textbf{38.53 \std{0.27}} & 35.56 \std{1.24} & 27.72 \std{1.81} & 22.68 \std{1.15} \\
CIFAR-100 & 68.07 \std{0.86} & 60.79 \std{0.40} & 68.69 \std{0.40} & \underline{75.67} \std{0.14} & \textbf{76.53 \std{0.05}} & 73.51 \std{0.41} & 73.54 \std{0.27} \\
MINC-2500 & 64.13 \std{0.09} & 63.10 \std{0.30} & 63.97 \std{0.14} & \underline{67.55} \std{1.09} & \textbf{68.48 \std{0.72}} & 67.19 \std{0.38} & 67.36 \std{0.31} \\
Food-101 & 69.41 \std{0.47} & 68.90 \std{0.77} & 70.74 \std{0.12} & 78.95 \std{0.34} & \textbf{79.59 \std{0.49}} & 78.66 \std{0.68} & \underline{78.98} \std{0.57} \\
PCam & 83.18 \std{0.42} & 82.47 \std{0.63} & 78.55 \std{2.05} & \underline{85.02} \std{0.61} & 83.82 \std{2.66} & 84.43 \std{0.41} & \textbf{85.41 \std{1.91}} \\
\bottomrule
\end{tabular}

\end{table}

\begin{figure}[hbpt]
  \centering
  \includegraphics[width=\linewidth]{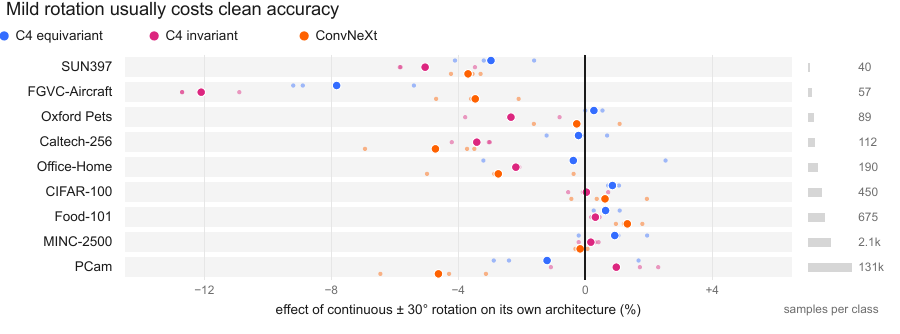}
  \caption{Impact of continuous $\pm 30^\circ$ rotation augmentation across model arms relative to unaugmented baselines, ordered by training set size.}
  \label{fig:mild}
\end{figure}

Our approach purposefully uses $C_4$-equivariance and compares against $C_4$ augmentations due to their lossless interpolation properties. 
In practice, rotations within a bounded range of the original orientation of the input image are common to mimic slight variations often occurring in natural images.
In addition to our previous results, we thus also include $\pm 30^\circ$ continuous rotation augmentation for completeness in Table~\ref{tab:panel_full}.
The table shows the validation macro-accuracy across all seven model arms and nine datasets.
On standard ConvNeXt, continuous $\pm 30^\circ$ augmentation reduces accuracy by \ensuremath{-1.97} points across 7/9 datasets.
On ConvNeXt-C4 Equivariant, $\pm 30^\circ$ augmentation is the only augmentation setting that improves performance on any dataset, achieving positive gains on 4/9 datasets with a modest average panel cost of \ensuremath{-1.10} points.
In comparison, ConvNeXt-C4 with invariant readout only benefits from continuous augmentation when dataset size increases (\textit{see Figure~\ref{fig:mild}}).

\subsection{Transformer Backbones: ReViT-C4}
\label{app:vit}

To test whether our findings extend beyond convolutional backbones, we additionally evaluate a vision transformer that efficiently encodes rotation equivariance.
ReViT~\citep{zaheer2026revit,zaheer2026revitv2} builds on the Convolutional vision Transformer~(CvT; \citealp{wu2021cvt}), which replaces linear token projections with convolutional attention projections and lifts these projections to group-convolutional self-attention over regular $C_4$ fields.
We compare CvT-13 with a $C_4$ adaptation of the tiny ReViT-v2 that uses the same equivariant orientation readout as ConvNeXt-C4 (Section~\ref{sec:method}), training both from scratch on the same datasets, splits, and epoch budgets as in Section~\ref{sec:regime}.
ReViT-C4 requires 1.7$\times$ fewer FLOPs than CvT-13, a reduction similar to that of ConvNeXt-C4 over ConvNeXt (1.6$\times$), while its parameter reduction is larger (9.5$\times$ versus 6.0$\times$).

\begin{wraptable}{r}{0.5\textwidth}
  \centering
  \caption{Validation accuracy (mean $\pm$ sample standard deviation across three seeds) from scratch, ordered by training set size. Bold is better.}
  \label{tab:revit_panel}
  \footnotesize
  \begin{tabular}{lrr}
\toprule
Dataset & CvT-13 & ReViT-C4 \\
\midrule
Oxford Pets & 53.06 \std{3.07} & \textbf{71.53 \std{3.34}} \\
FGVC-Aircraft & 39.10 \std{1.25} & \textbf{41.70 \std{5.20}} \\
Office-Home & 50.79 \std{0.23} & \textbf{62.64 \std{0.70}} \\
Caltech-256 & 46.49 \std{0.79} & \textbf{67.61 \std{0.30}} \\
SUN397 & 26.79 \std{1.21} & \textbf{40.56 \std{0.52}} \\
CIFAR-100 & 66.77 \std{0.87} & \textbf{76.22 \std{0.11}} \\
MINC-2500 & 62.10 \std{0.37} & \textbf{70.72 \std{0.41}} \\
Food-101 & 68.41 \std{0.45} & \textbf{77.19 \std{0.21}} \\
PCam & 88.49 \std{0.96} & \textbf{88.67 \std{0.89}} \\
\bottomrule
\end{tabular}

\end{wraptable}

\begin{figure}[hbpt]
  \centering
  \includegraphics[width=\linewidth]{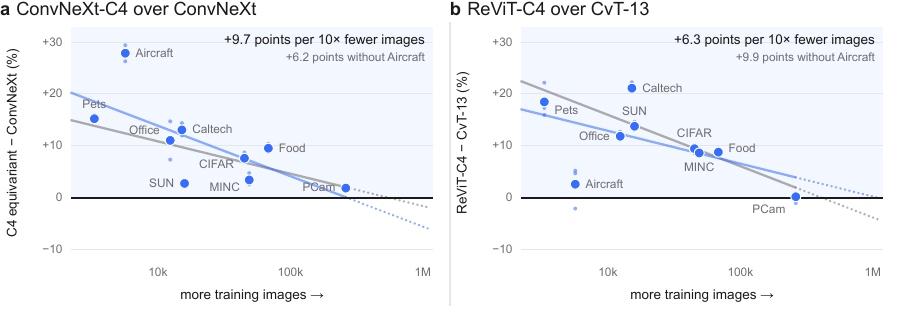}
  \caption{\textbf{Data-regime dependency for convolution and transformer backbones.} From-scratch validation macro-accuracy gain of the $C_4$-equivariant model over its unconstrained baseline against training set size for \textbf{(a)} ConvNeXt-C4 over ConvNeXt and \textbf{(b)} ReViT-C4 over CvT-13. Small dots show seeds, large dots the 3-seed mean. The blue line fits all nine dataset means against $\log_{10}$ training set size; the grey line repeats the fit without FGVC-Aircraft.}
  \label{fig:revit_regime}
\end{figure}

From scratch, ReViT-C4 outperforms CvT-13 on all nine datasets (Table~\ref{tab:revit_panel}) by \ensuremath{+10.54} points on average.
The accuracy advantage is largest in the low-data regime, mirroring the trend for ConvNeXt-C4 and reaching \ensuremath{+21.12} points on Caltech-256, \ensuremath{+18.47} points on Oxford Pets, while diminishing to \ensuremath{+0.18} points on PCam (Figure~\ref{fig:revit_regime}b).
The ordering within the datasets differs slightly from ConvNeXt-C4: across all nine datasets, the ReViT-C4 gain drops by 6.33 points per tenfold increase in training set size (Spearman $\rho = \ensuremath{-0.53}, p = 0.148$), compared to 9.72 points for ConvNeXt-C4 ($\rho = \ensuremath{-0.82}, p = 0.011$).
This is possibly due to the larger architectural gap between CvT-13 and ReViT-C4: beyond the symmetry constraint, ReViT also changes the number of stages, replaces global with windowed attention, and uses different normalization layers and nonlinearities, whereas ConvNeXt-C4 stays closer to its non-equivariant baseline.

The FGVC-Aircraft dataset shows interesting differences between the two architecture families.
While ConvNeXt-C4 achieves its largest gain on this dataset (\ensuremath{+27.87} points), ReViT-C4 improves over CvT-13 by only \ensuremath{+2.60} points and has the highest seed variance in Table~\ref{tab:revit_panel}.
CvT-13 already reaches 39.10\% compared to 29.13\% for ConvNeXt, while ReViT-C4 reaches 41.70\% compared to 57.00\% for ConvNeXt-C4.
Dataset properties other than size thus still determine how much an architectural inductive bias helps, at least between transformer- and convolution-based backbones, which is an interesting direction for future research.
Excluding FGVC-Aircraft, both comparisons follow a similar log-linear trend (grey lines in Figure~\ref{fig:revit_regime}): the accuracy gain changes to 9.90 points per tenfold increase in training set size for ReViT-C4 ($\rho = \ensuremath{-0.86}, p = 0.011$) and to 6.23 points for ConvNeXt-C4 ($\rho = \ensuremath{-0.76}, p = 0.037$).
Overall, these results corroborate Finding~2 across convolutional and transformer architectures: equivariance compensates for data scarcity.

\subsection{Natural Distribution Shifts and Robustness Evaluations}
\label{app:shift}

\begin{figure}[hbpt]
  \centering
  \includegraphics[width=\linewidth]{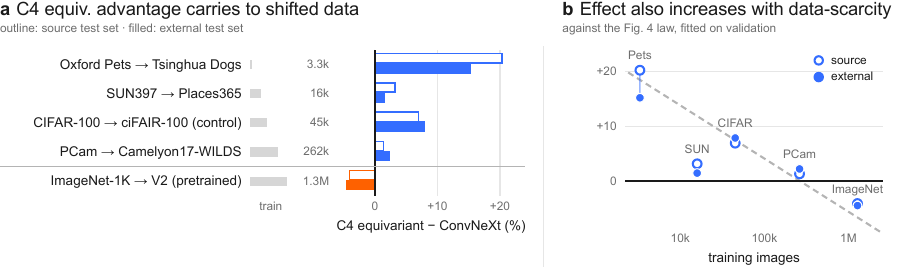}
  \caption{\textbf{Evaluation under natural distribution shifts.} \textbf{(a)} Accuracy differences ($\text{C4} - \text{ConvNeXt}$) on source test sets (outline) and external shifted test sets (filled), ordered by training set size. \textbf{(b)} Shift accuracy differences plotted against training set size, overlaid with the from-scratch validation scaling law (dashed line).}
  \label{fig:shift}
\end{figure}

To assess whether the representations learned by ConvNeXt-C4 exhibit out-of-distribution robustness, we evaluate models on four natural distribution shift benchmarks (Figure~\ref{fig:shift}):
(1)~\textbf{Oxford Pets~$\to$~Tsinghua Dogs:} Evaluates fine-grained animal category transfer under geographic and camera shift.
We purged 501 byte-exact duplicate images between Oxford Pets and Tsinghua Dogs.
ConvNeXt-C4 leads by \ensuremath{+15.26} points.
(2)~\textbf{PCam~$\to$~Camelyon17-WILDS:} Evaluates histopathology metastasis detection across 10 held-out hospital slide centers.
Our ConvNeXt-C4 outperforms ConvNeXt by \ensuremath{+2.25} points overall and leads on 9 of 10 individual hospital slides (maximum gain of \ensuremath{+8.81} points).
(3)~\textbf{SUN397~$\to$~Places365:} Evaluates scene category classification under cross-dataset collection shift.
ConvNeXt-C4 leads by \ensuremath{+1.49} points.
(4)~\textbf{ImageNet-1K~$\to$~ImageNet-V2:} Evaluates large-scale transfer under dataset replication shift.
The pre-trained ConvNeXt baseline leads by \ensuremath{4.45} points.

\subsection{Orientation-Selection Classifier Two-Sample Test}
\label{app:diagnostics}

Our orientation-selection classifier two-sample test is defined as following:
Let $P(x)$ denote an empirical image distribution and $\bar{P}(x) = \frac{1}{4}\sum_{t=0}^3 R_t \# P(x)$ the $C_4$-balanced mixture.
We use the \textit{C2ST-NN} from \citet{lopezpaz2017revisiting} as the classifier to discriminate between $P(x)$ and $\bar{P}(x)$ over three detector seeds.
The model has a single hidden layer with ReLU non-linearities and is trained on our binary classification task.
For all datasets, images are resized and where necessary center-cropped to $64\times64$ pixels.
The global orientation-selection accuracy uses these full images, while the local score is based on randomly sampled $32\times32$ patches.
An accuracy of $50\%$ in this test indicates that the dataset exhibits perfect rotational symmetry whereas $87.5\%$ indicates that all three orientations are completely distinct with respect to the base image.

\subsection{Representation Probes and Spatial Feature Analysis}
\label{app:probes}

\begin{figure}[hbpt]
  \centering
  \includegraphics[width=\linewidth]{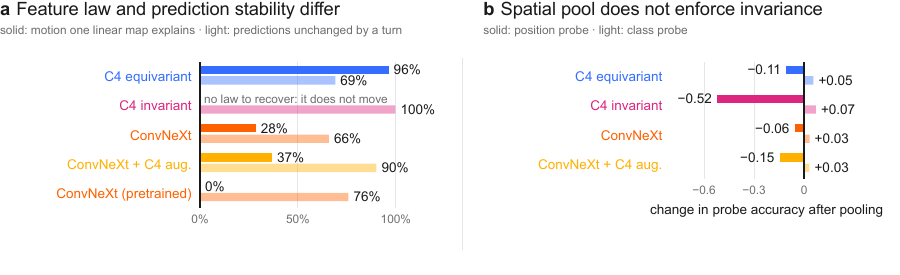}
  \caption{\textbf{Feature representation probes on frozen encoders.} \textbf{(a)} Percentage of feature transformation explained by a linear group action vs.\ prediction stability under quarter-turns. \textbf{(b)} Performance change in linear probes for spatial position and class semantics before and after global average pooling.}
  \label{fig:probes}
\end{figure}

Figure~\ref{fig:probes} analyzes geometric properties of intermediate representations:\newline
(1)~\textbf{Action Law Recovery vs.\ Output Stability:} For ConvNeXt-C4 with equivariant readout, a linear map can explain 96\% of feature variance under quarter-turns, while 69\% of predictions remain unchanged.
Standard ConvNeXt explains 28\% of feature variance and retains 66\% prediction stability.
Augmenting standard ConvNeXt increases prediction stability to 90\% but only explains 37\% of feature transformation, illustrating that augmentation enforces behavioral invariance without structuring internal representations.\newline
(2)~\textbf{Spatial Global Average Pooling:} As our ConvNeXt-based models are also translation equivariant, we briefly evaluate if retaining spatial information has a similar benefit to keeping orientation information for our $C_4$ model.
Figure~\ref{fig:probes}b shows that position information is lost across all four evaluated models when applying spatial pooling.
Notably, applying $C_4$ augmentations and the invariant $C_4$ readout both lead to worse preservation of positional information.
However, when evaluating linear probing before and after global average pooling, we see that spatial pooling improves semantic class probing accuracy across all setups.
This indicates, that positional information might be less relevant for classification compared to other tasks such as vision-language modeling, where spatial information is commonly preserved~\citep{liu2023llava,steiner2024paligemma}.
Future work could evaluate if preserving orientation information might thus be even more relevant in tasks other than image classification.

\end{document}